\documentclass{article}
\usepackage{spconf,amsmath,graphicx}
\usepackage{amsfonts}
\usepackage{amssymb}
\usepackage{amsmath}
\DeclareMathOperator*{\argmax}{argmax}
\usepackage{hhline, booktabs}
\usepackage{multirow}
\usepackage{tikz}
\usepackage{subcaption}
\usepackage{makecell}
\definecolor{lavender}{rgb}{0.97,0.97,0.97}
\definecolor{best}{rgb}{0.24, 0.7, 0.44}
\definecolor{second_best}{rgb}{0.82, 0.94, 0.75}
\usepackage{xcolor,colortbl}
\usepackage{ctable}
\usepackage{comment}
\usepackage[absolute,overlay]{textpos}
\usepackage{MnSymbol}
\usepackage{xcolor}
\usepackage[normalem]{ulem}

\usepackage{enumitem}
\setlist{nosep, leftmargin=14pt}

\usepackage{mwe} 

\title{Towards Concept-based Interpretability of Skin Lesion Diagnosis using Vision-Language Models}
\name{Cristiano Patrício$^{1,3}$, Luis F. Teixeira$^{2,3}$, João C. Neves$^{1}$}
\address{$^1$Universidade da Beira Interior and NOVA LINCS, \\ $^2$Faculdade de Engenharia da Universidade do Porto, $^3$INESC TEC}
\begin{document}
%
\maketitle
\begin{abstract}
Concept-based models naturally lend themselves to the development of inherently interpretable skin lesion diagnosis, as medical experts make decisions based on a set of visual patterns of the lesion. Nevertheless, the development of these models depends on the existence of concept-annotated datasets, whose availability is scarce due to the specialized knowledge and expertise required in the annotation process. In this work, we show that vision-language models can be used to alleviate the dependence on a large number of concept-annotated samples. In particular, we propose an embedding learning strategy to adapt CLIP to the downstream task of skin lesion classification using concept-based descriptions as textual embeddings. Our experiments reveal that vision-language models not only attain better accuracy when using concepts as textual embeddings, but also require a smaller number of concept-annotated samples to attain comparable performance to approaches specifically devised for automatic concept generation.
\end{abstract}
\section{Introduction}
\label{sec:intro}

Automated computer-aided diagnosis systems for disease detection from medical images have undergone a remarkable increase in performance, primarily attributed to the enhanced capabilities of deep learning models. This paradigm has led to a substantial increase in the precision of these systems in providing accurate diagnosis in various medical image tasks, such as skin lesion diagnosis, assuring in some cases results that match the performance of dermatologists~\cite{codella2017deep, esteva2017dermatologist}. However,  the ``black-box'' nature of these deep learning-based systems in dermatology poses the most significant barrier to their broad adoption and integration into clinical workflow~\cite{rotemberg2019role}. 
To alleviate this problem, interpretability methods have emerged to ensure the transparency and robustness of medical AI systems. Among these interpretable strategies, Concept Bottleneck Models (CBM)~\cite{koh_2020_ICML} are growing in popularity in medical imaging analysis~\cite{Fang_MM2020,Lucieri_IJCNN2020,Patricio_2023_CVPRW}, since they allow to explain the decision process based on the presence or absence of human-understandable concepts, which aligns perfectly with the way clinicians draw conclusion from medical images. Furthermore, several studies concluded that humans prefer concept-based explanations over other forms of explanations, such as heatmaps or example-based~\cite{ramaswamy2023overlooked}. In spite of their popularity, the development of concept-based models depends on dense annotations of human-understandable concepts~\cite{yang2023language}, which are time-consuming and require expertise from domain experts, limiting the adoption of such models in medical image tasks. Several works~\cite{yang2023language,oikarinen2023label,menon2022visual,yan2023robust} attempt to mitigate this problem by querying Large Language Models (LLMs) to generate additional information about target classes to form candidate concepts. 

In this work, we show that despite these advances, detailed concept-based descriptions generated from LLMs lead to inferior classification accuracy when compared with the use of textual embeddings derived directly from dermatoscopic concepts. Specifically, we compared the performance of LLMs on three well-known skin lesion datasets~\cite{PH2, DERM7PT, codella2019skin} using three distinct strategies for measuring the similarity between a given query skin image and textual embeddings: (i) utilizing the target class as textual embedding; (ii) using a set of dermoscopic concepts annotated by board-certified dermatologists as textual embeddings; and (iii) leveraging concept descriptions generated by ChatGPT as textual embeddings.
Our experiments reveal that (i) relying on expert-selected dermoscopic concepts as textual embeddings leads to better performance in distinguishing melanoma from other diseases, in addition to providing concept-based explanations and (ii) a simple and efficient embedding learning procedure on top of feature embeddings of CLIP~\cite{radford2021learning} could attain comparable performance to models specifically designed for the task of automated concept generation of dermoscopic features.

Our contributions can be summarized as follows: (i) we introduce an efficient and simple embedding learning procedure to improve the performance of CLIP models in the downstream task of melanoma diagnosis; (ii) we alleviate the annotation burden of CBMs by using zero-shot capabilities of Vision Language Models (VLMs) to automatically annotate concepts; (iii) we provide concept-based explanations for the model prediction based on expert-selected dermoscopic concepts.

\begin{figure*}[t]
    \centering
    \includegraphics[width=\textwidth]{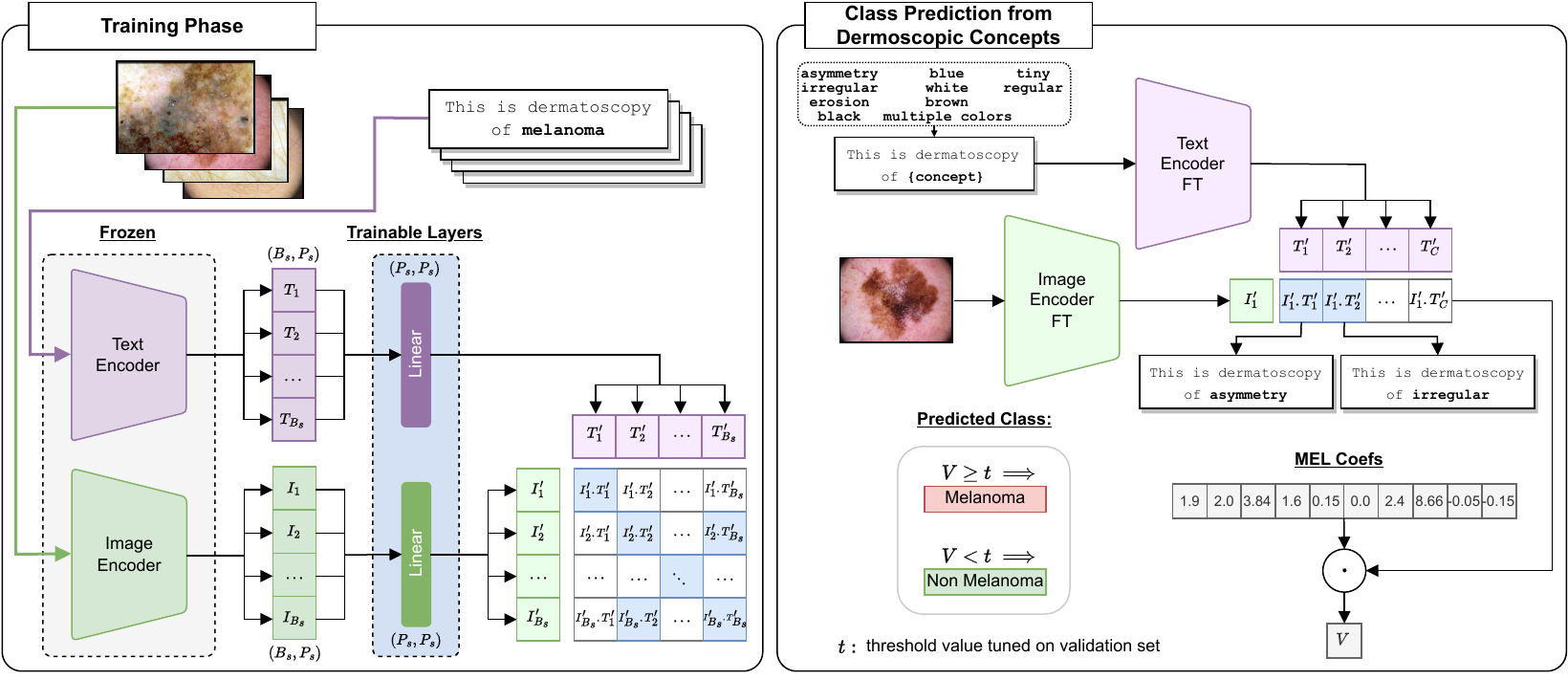}
    \caption{\textbf{The workflow of our proposed strategy}. After learning the new multi-modal embedding space (left), we predict the presence of melanoma by linearly combining the similarity scores with the melanoma coefficients acting as the bottleneck layer of CBM. The result of this operation is then compared with a threshold value to predict the presence or absence of melanoma.}
    \label{fig:method}
\vspace{-0.3cm}
\end{figure*}

\section{Method}
\label{sec:method}

Figure \ref{fig:method} presents an overview of the proposed method. The training phase consists of learning a new multi-modal embedding space for approximating image and textual embeddings of the same category (section \ref{subsec:ft_clip}). The learned projection layers are then used to calculate the feature embeddings of both the image and textual descriptions in order to predict melanoma (section \ref{subsec:melanoma_prediction}) by: (i) calculating the cosine similarity between the image feature and the text encoding of each disease label; (ii) calculating the cosine similarity between the image and a concept $c$ in the concept set $C$, whose scores are then fed into the classification layer to determine the presence of melanoma; or (iii) calculating the cosine similarity between the image and a set of $m$ concept descriptors per concept $c$, average the scores per concept, and then fed into the classification layer as in (ii).

\subsection{Embedding Learning}
\label{subsec:ft_clip}

Let $\mathcal{D} = \{(i,y)\}$ be a batch of image-label pairs where $i$ is the image and $y \in \mathcal{Y}$, is a label from a set of $N$ classes. We extract the features of the frozen CLIP image encoder $\mathcal{I}(.)$ and the text encoder $\mathcal{T}(.)$ to obtain the feature embedding of the image $x = \mathcal{I}(i) \in \mathbb{R}^d$ and the feature embedding of the label $l = \mathcal{T}(y) \in \mathbb{R}^d$. The training phase (Figure \ref{fig:method}) thus consists of learning a new multi-modal embedding space by jointly training an image projection layer $W_I$ and text projection layer $W_T$ to maximize the cosine similarity of the image feature $W_I . x$ and text feature $W_T.l$ embeddings of the $n$ pairs sharing the same disease while minimizing the cosine similarity of embeddings of the pairs from different diseases. For this, we define a target matrix as having ones on image-label pairs sharing the same disease label, and zeros in the remaining pairs. We adopt the objective function used in~\cite{radford2021learning}.

\vspace{-0.2cm}
\subsection{Strategies for Melanoma Diagnosis}
\label{subsec:melanoma_prediction}

\textbf{Baseline}\hspace{0.3cm}
\sloppy The most straightforward strategy for using CLIP in the task of melanoma classification is to calculate the similarity between the visual descriptor of the image $x = \mathcal{I}(i)$ and the textual feature representation of the $N$ disease labels $l = \mathcal{T}(y), \; y \in \mathcal{Y}$. The predicted disease label is given by $\hat{y} = \argmax_{y \in \mathcal{Y}} \;S_c(W_I.x, W_T.l)$, where $S_c$ is the cosine similarity.\\

\vspace{-0.2cm}
\noindent \textbf{CBM}\hspace{0.3cm}
Alternatively, we can calculate the degree to which a dermoscopic concept $c \in C = \{c_1,...,c_{N_c}\}$ is present in the image by measuring the similarity between the feature embedding of the image $x = \mathcal{I}(i)$ and each feature embedding of concept $c$ given by $E_C \in \mathbb{R}^{N_C \times d}$, where each row of $E_C$ is a text feature $\mathcal{T}(c) \in \mathbb{R}^d$ of a concept $c$. Then, we employ the dermoscopic concept coefficients (\underline{\textbf{MEL Coefs}} in Figure \ref{fig:method}) extracted from a previously trained linear model for melanoma prediction~\cite{kim2023fostering}, denoted as $\mathcal{W}_{mel} \in \mathbb{R}^{1 \times N_C}$, and multiply them with the obtained concept scores $p = S_c(W_I.x,W_T.E_C), \; p \in \mathbb{R}^{N_C \times 1}$. Let $V = \mathcal{W}_{mel} \cdot p$. The final prediction is thus given by $\hat{y} = \left\{\begin{matrix}0, & \mathrm{if} \; V < t \\1, & \mathrm{if} \; V \geq t \end{matrix}\right.$, where $t$ is a threshold value tuned on the validation set.

\vspace{0.2cm}
\noindent \textbf{GPT + CBM}\hspace{0.3cm}
We query ChatGPT with a designed prompt to generate a set of $m$ textual descriptions for a given dermoscopic concept $c$. The chosen prompt ``According to published literature in dermatology, which phrases best describe a skin image containing \{\texttt{concept}\}?'' returns a total of five descriptions for each individual concept $c$ (see supplementary). We obtain the feature embedding for the $m$ descriptions $E_{s^c} = \mathcal{T}(s^c_1, ..., s^c_m),\; E_{s^c} \in \mathbb{R}^{m \times d}$ of a concept $c$. We calculate the concept scores as $p_c = \frac{1}{m} \sum_{i=0}^{m} S_c(W_I.x,W_T.E_{s_i^c})$. Let $V = \mathcal{W}_{mel} \cdot \sum_{c=0}^{N_c} p_c$. The final score indicating the presence of melanoma is thus given by $\hat{y} = \left\{\begin{matrix}0, & \mathrm{if} \; V < t \\1, & \mathrm{if} \; V \geq t \end{matrix}\right.$.

\section{Experimental Setup}
\label{sec:experimental_results}

We evaluate different CLIP variations, using our proposed embedding learning, and compare it with MONET~\cite{kim2023fostering}, a foundation model trained on dermatological images, under the previously defined strategies (section \ref{subsec:melanoma_prediction}) on three dermoscopic datasets. Also, we report the performance of a black-box linear probing model to assess whether our approach can maintain black-box accuracy without compromising interpretability.

\vspace{0.2cm}
\noindent \textbf{Datasets}\hspace{0.3cm} 
Three dermoscopic datasets were selected for our experiments, namely: PH$^2$~\cite{PH2}, Derm7pt~\cite{DERM7PT} and ISIC 2018~\cite{codella2019skin}. The PH$^2$ dataset encompasses dermoscopic images of melanocytic lesions, including ``melanoma'' and two types of ``nevus'' that were merged and treated as singular ``nevus''. For PH$^2$, we used 5-fold cross-validation. Derm7pt comprises clinical and dermoscopic images, which we filtered to obtain images of ``nevus'' and ``melanoma'' classes. ISIC 2018 is composed of dermoscopic images including different types of skin lesions, namely ``melanoma'', ``melanocytic nevus'', ``basal cell carcinoma'', ``actinic keratosis'', ``benign keratosis'', ``dermatofibroma'', and ``vascular lesion''. Detailed statistics of the datasets, including the train/val/test splits, are presented in Table \ref{tab:dataset_statistics}\footnote{We followed the split partition adopted in~\cite{DERM7PT} for the Derm7pt dataset and in~\cite{barata2023reinforcement} for ISIC 2018.}.

\begin{table}[!ht]
  \centering
  \setlength{\tabcolsep}{5pt}
  \resizebox{0.45\textwidth}{!}{%
  \begin{tabular}{lcccc}
    \toprule
     \textbf{Dataset} & \textbf{Classes} & \textbf{Train size} & \textbf{Validation size} & \textbf{Test size} \\
    \midrule
    PH$^2$~\cite{PH2} & 2 & 160 (28 to 34) & - & 40 (6 to 12)\\
    Derm7pt~\cite{DERM7PT} & 2 & 346 (90) & 161 (61) & 320 (101) \\
    ISIC 2018~\cite{codella2019skin} & 7 & 8,012 (890) & 2,003 (223) & 1,511 (171)\\
    \bottomrule
  \end{tabular}%
  }
  \caption{\textbf{Dataset statistics}. Numbers between rounded brackets represent the $\#$ of Melanoma examples in the split.}
  \label{tab:dataset_statistics}
\vspace{-0.6cm}
\end{table}

\subsection{Implementation Details}

\noindent \textbf{Embedding Learning}\hspace{0.3cm} 
The projection layers (section \ref{subsec:ft_clip}) were trained on Derm7pt and ISIC 2018 datasets using the AdamW optimizer with a learning rate of $1e^{-5}$. Also, a learning rate decrease policy was used with a patience of $1$ and a factor of $0.8$.
The trainable projection layers are linear layers with the same dimension of the output of image and text encoder of CLIP\footnote{The source code and supplementary material are available at {https://github.com/CristianoPatricio/concept-based-interpretability-VLM}}. For the evaluation of MONET we follow the proposed strategy by the authors to calculate the concept scores. For the black-box linear probing, we follow~\cite{radford2021learning} and use image features taken from the penultimate layer of each model, ignoring any classification layer provided. A logistic regression classifier is trained on the top of the extracted image features using scikit-learn's L-BFGS implementation, with maximum $1,000$ iterations. 

\noindent \textbf{Preprocessing}\hspace{0.3cm} The input images were preprocessed according to the transformations defined in the original image encoders of CLIP variations. Additionally, and following~\cite{Patricio_2023_CVPRW}, we use segmented versions of the images. This strategy ensures that solely the area of the lesion is considered, preventing the model from giving attention to artifacts in the image. Most importantly, this procedure allows improving the final classification results. 

\section{Results}
\label{sec:results}




\subsection{Quantitative Analysis}

\begin{figure}[h!]
\begin{minipage}[b]{.48\linewidth}
  \centering
  \centerline{\includegraphics[width=4.5cm]{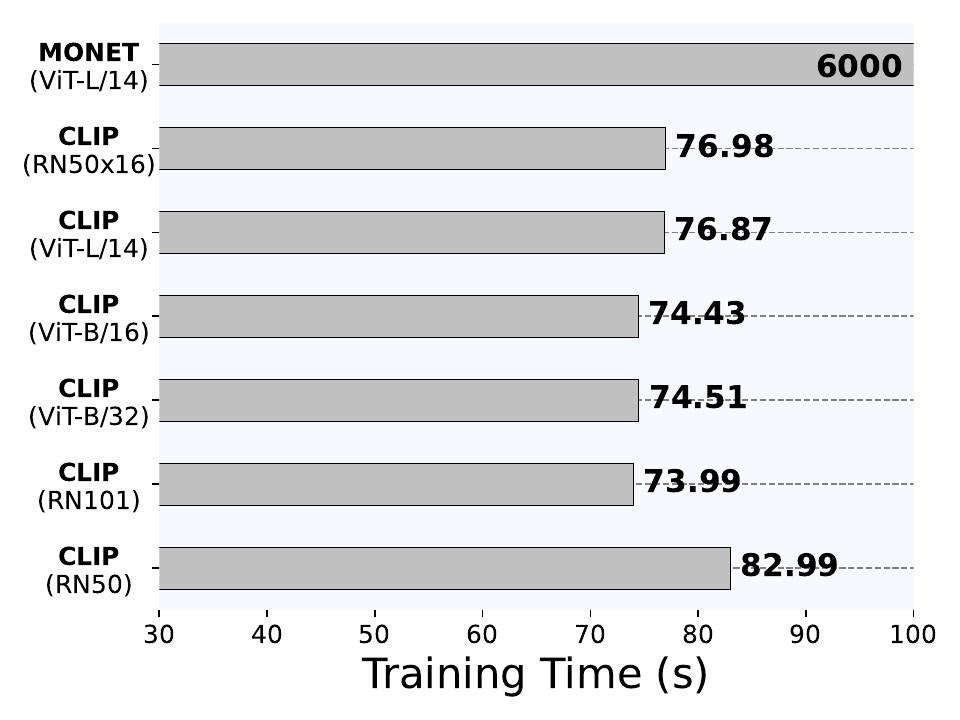}}
\end{minipage}
\hfill
\begin{minipage}[b]{0.48\linewidth}
  \centering
  \centerline{\includegraphics[width=4.8cm]{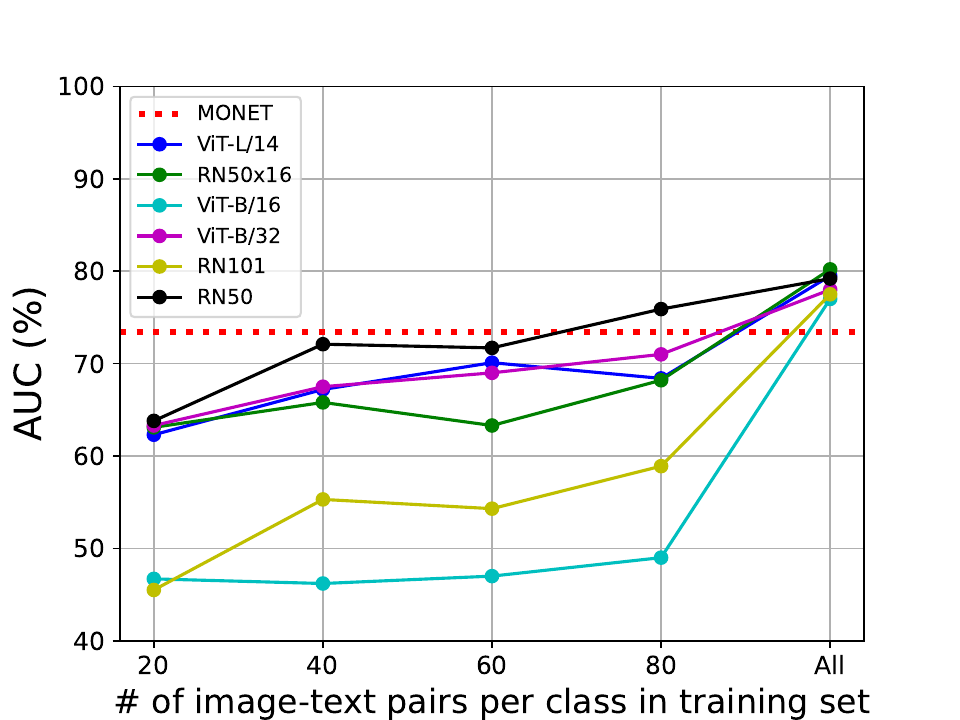}}
\end{minipage}
\caption{Computational performance analysis of our proposed embedding learning procedure.}
\label{fig:performance_comparison_ft}
\vspace{-0.3cm}
\end{figure}

\noindent \textbf{Comparison with Original CLIP and MONET}\hspace{0.3cm} Table \ref{tab:performance_clip_ft} compares the performance of the original CLIP model with our method across three different strategies on two datasets. The reported results represent the average Balanced Accuracy (BACC) obtained across CLIP model variations for each specific strategy. Our method outperforms CLIP original variations by an average of $11.5\%$ and $9.2\%$ on both datasets, respectively. The most significant improvement is observed on the Baseline strategy for Derm7pt, and on CBM strategy for ISIC 2018. Figure \ref{fig:performance_comparison_ft} (left) shows the efficiency of our method in comparison to the MONET model. Notably, our method achieves a comparable level of performance of MONET while requiring significantly less training time. On the other hand, Figure \ref{fig:performance_comparison_ft} (right) depicts the evolution of AUC (in \%) as more image-label pairs are added into the training set of ISIC 2018. The results show that CLIP RN50, CLIP ViT L/14 and CLIP ViT-B/32 attain comparable performance with MONET when using only between 40-60 image-label pairs in the training set. 

\begin{table}[h!]
  \centering
  \setlength{\tabcolsep}{5pt}
  \resizebox{0.37\textwidth}{!}{%
  \begin{tabular}{lcccc}
    \toprule
     \multirow{2}{*}{\textbf{Strategy}} & \multicolumn{2}{c}{\textbf{Derm7pt}~\cite{DERM7PT}} & \multicolumn{2}{c}{\textbf{ISIC 2018}~\cite{codella2019skin}} \\
     \cmidrule(lr){2-3} \cmidrule(lr){4-5}
      & Orig. & Ours & Orig. & Ours \\
    \midrule
    Baseline & $61.3 \pm 2.4$ & $\mathbf{75.0 \pm 2.5}$ & $54.1 \pm 5.0$ & $\mathbf{63.2 \pm 1.4}$ \\
    CBM & $65.4 \pm 2.6$ & $\mathbf{75.4 \pm 2.3}$ & $60.6 \pm 3.0$ & $\mathbf{70.4 \pm 3.0}$ \\
    GPT+CBM & $64.1 \pm 6.3$ & $\mathbf{74.9 \pm 2.6}$ & $61.2 \pm 3.2$ & $\mathbf{69.9 \pm 3.2}$ \\
    \bottomrule
  \end{tabular}%
  }
  \caption{Performance gains of CLIP with our proposed embedding learning strategy in terms of BACC.}
  \label{tab:performance_clip_ft}
\vspace{-0.3cm}
\end{table}

\begin{figure*}[!ht]
    \begin{tikzpicture}[font=\scriptsize]
        \node[anchor=west] (ph2) at (0,0) {\includegraphics[width=0.33\textwidth]{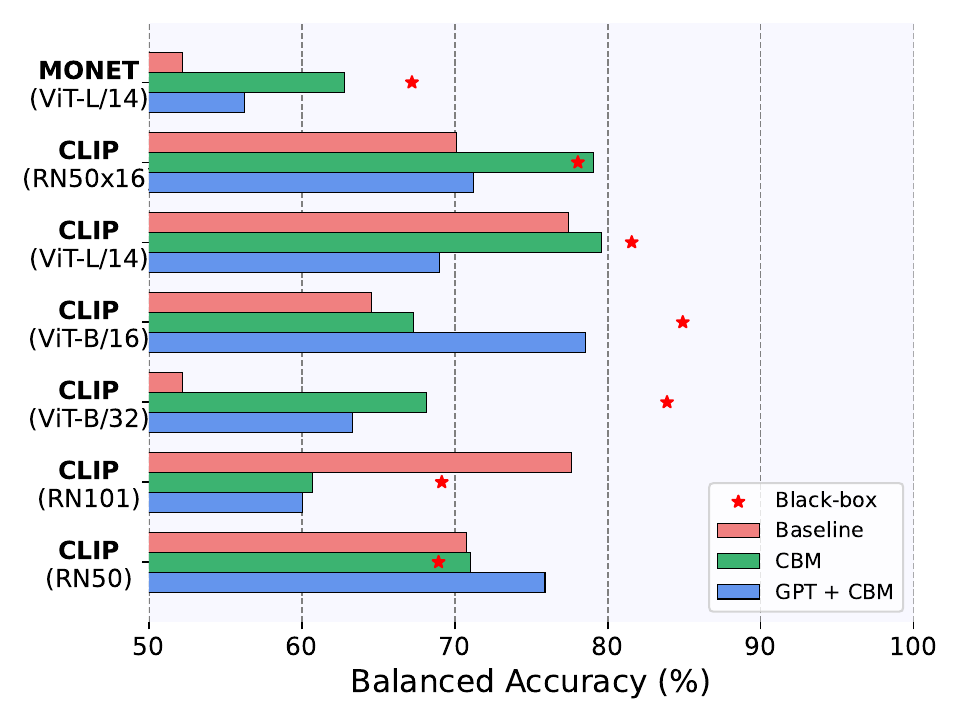}};

        \node[anchor=west] (derm7pt) at (0.33\textwidth,0) {\includegraphics[width=0.33\textwidth]{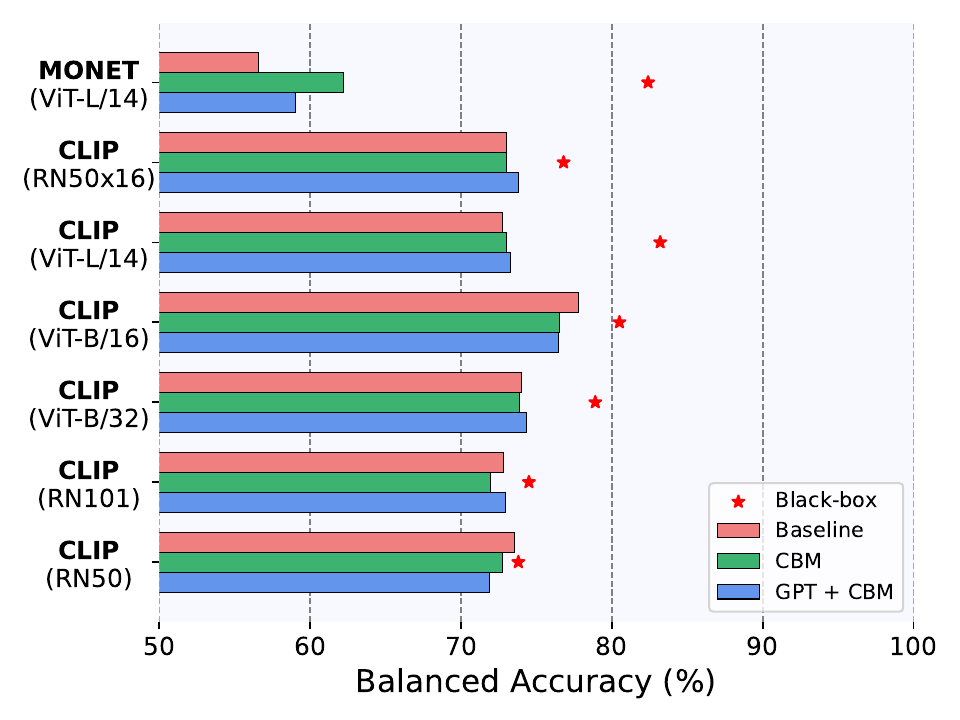}};

        \node[anchor=west] (isic) at (0.66\textwidth,0) {\includegraphics[width=0.33\textwidth]{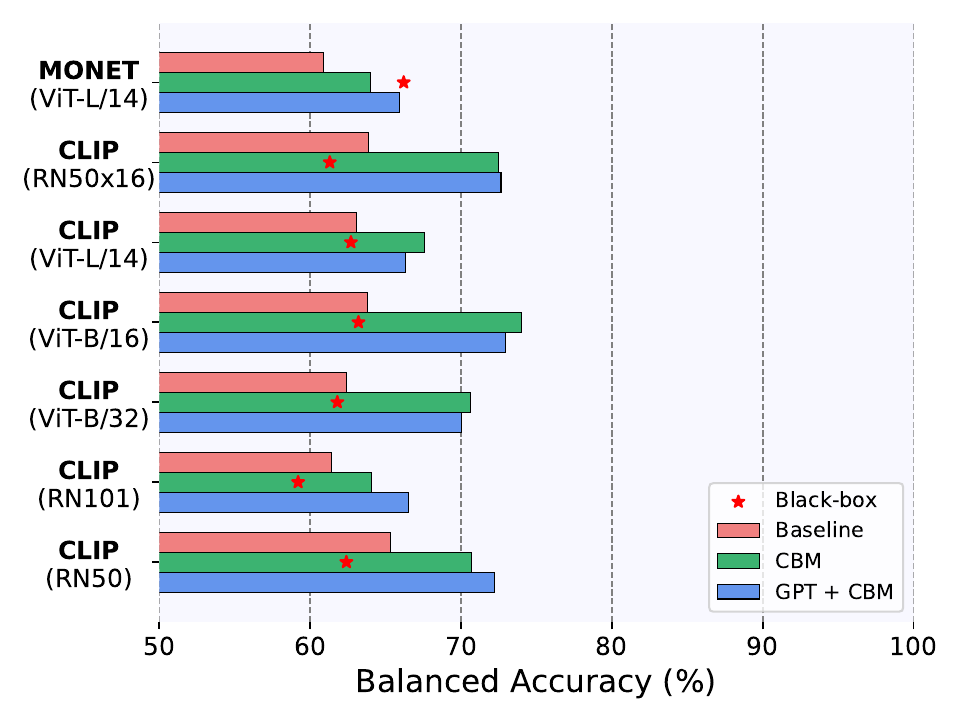}};

        \node at (0.175\textwidth,2.4) {\textbf{PH2}};
        \node at (0.52\textwidth,2.4) {\textbf{Derm7pt}};
        \node at (0.85\textwidth,2.4) {\textbf{ISIC 2018}};
    \end{tikzpicture}
    \caption{Evaluation results (in BACC \%) of the different classification strategies (Baseline, CBM and GPT+CBM) on three datasets (PH$^2$, Derm7pt and ISIC 2018) for melanoma detection. Black-box linear probing performance is marked with \textcolor{red}{$\filledstar$}.}
    \label{fig:results}
\end{figure*}
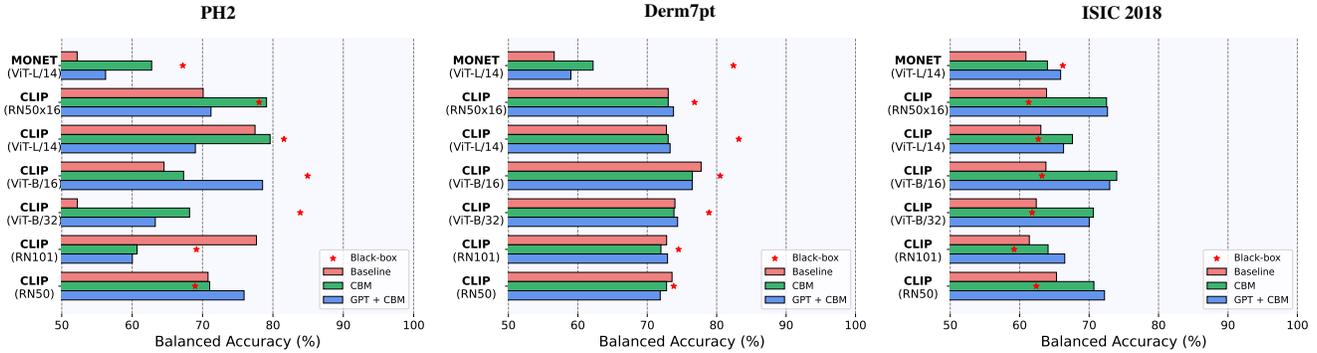

\noindent \textbf{Evaluation of different VLMs for melanoma diagnosis}\hspace{0.3cm} 
The results presented in Figure \ref{fig:results} show the performance in terms of BACC. For the PH$^2$ dataset, the results represent the average performance over 5-fold cross-validation. The results reported for Derm7pt and ISIC 2018 datasets are the averages obtained from four separate runs. The results on PH$^2$ dataset suggest that the GPT-CBM strategy outperforms both the Baseline and CBM strategies for CLIP ViT-B/16. Additionally, the CBM strategy demonstrates statistically significant improvement over the GPT+CBM strategy when applied to RN50x16.
Regarding the Derm7pt dataset, all strategies exhibit comparable performance. However, a marginal gain of GPT+CBM over CBM and the Baseline is noticeable in 4 out of 7 models. In the case of ISIC 2018, the results show significant improvement of both CBM and GPT+CBM strategies over the Baseline ($p < 0.05$).

\subsection{Interpretability by Dermoscopic Concepts}
\begin{figure}[!ht]
\begin{minipage}[b]{\linewidth}
  \centering
  \centerline{\includegraphics[width=0.8\textwidth,height=3.5cm]{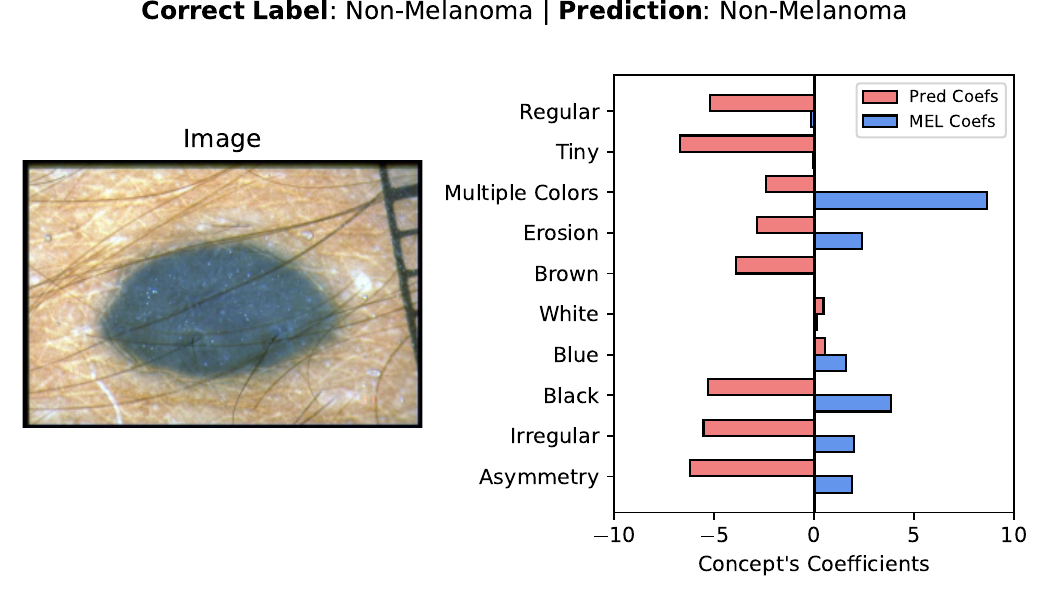}}
\end{minipage}
\hfill
\begin{minipage}[b]{\linewidth}
  \centering
  \centerline{\includegraphics[width=0.8\textwidth,height=3.5cm]{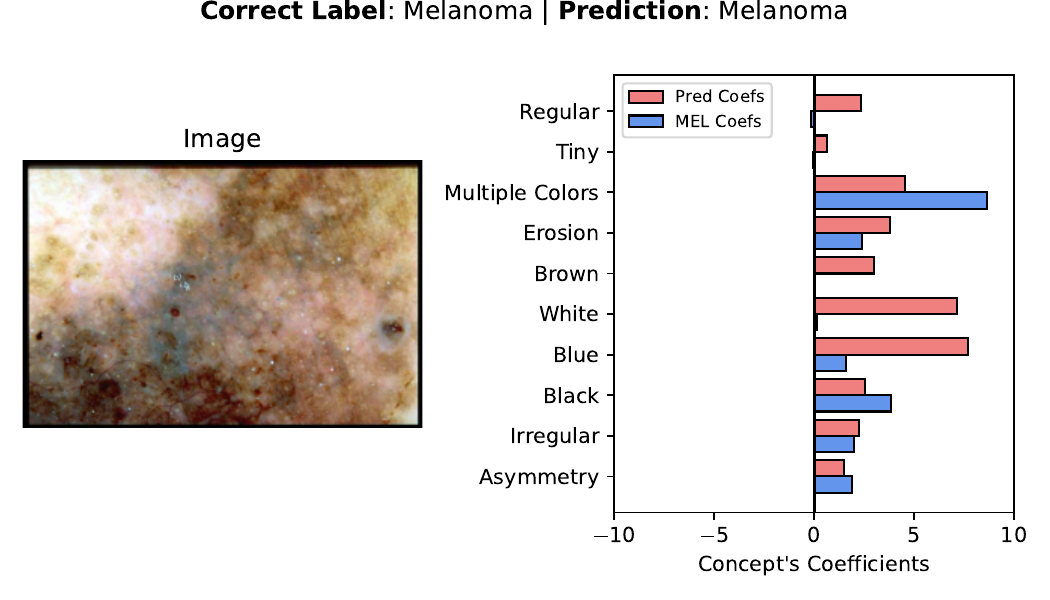}}
\end{minipage}
\caption{Examples of dermoscopic images classified based on dermoscopic concepts.}
\label{fig:qualitative_results}
\vspace{-0.5cm}
\end{figure}

Utilizing dermoscopic concepts for melanoma detection ensures the interpretability and transparency of the model's decision-making process. In Figure \ref{fig:qualitative_results}, we present two illustrative examples, each accompanied by the predicted dermoscopic concepts. In the upper image, the model classifies it as non-melanoma, as indicated by the negative contributions of dermoscopic concepts typically associated with melanoma. Conversely, the lower image was correctly classified as melanoma, as evidenced by the positive contributions of melanoma-specific concepts, which align with the ABCDEs of melanoma~\cite{rigel2005abcde}. Additional examples can be found in the supplementary material.

\section{Conclusions and Future Work}
\label{sec:conclusion}
This paper presents an efficient embedding learning procedure to enhance the performance of CLIP models in the downstream task of melanoma diagnosis, utilizing various strategies. Our comparative evaluation of VLMs' efficacy in melanoma diagnosis indicates that predicting melanoma based on expert-selected dermoscopic concepts is more reliable than using the textual description of the target class, promoting interpretability in decision-making. Additionally, our experiments suggest that incorporating detailed descriptions of concepts as a proxy to use them directly in predicting melanoma does not lead to statistically significant improvements. In future research, we plan to expand the analysis to other imaging modalities to foster trust and acceptance of automated diagnosis systems in daily clinical practices.

\vfill
\pagebreak

\noindent \textbf{Acknowledgments}\hspace{0.3cm} This work was funded by the Portuguese Foundation for Science and Technology (FCT) under the PhD grant ``2022.11566.BD'', and supported by NOVA LINCS (UIDB/04516/2020) with the financial support of FCT.IP.

\vspace{0.3cm}

\noindent \textbf{Compliance with ethical standards}\hspace{0.3cm} This research study was conducted using human subject data, available in open access. Ethical approval was not required.

\small{
\bibliographystyle{IEEEbib}
\bibliography{strings,refs}
}

\end{document}